\documentclass[runningheads]{llncs}
\usepackage[T1]{fontenc}
\usepackage{graphicx}
\usepackage{hyperref}
\usepackage{color}

\begin{document}
\title{Prediction of Tourism Flow with Sparse Geolocation Data}
%
%
\author{Julian Lemmel*\inst{1,2} \and
Zahra Babaiee*\inst{1,2} \and
Marvin Kleinlehner\inst{1} \and Ivan Majic\inst{3} \and Philipp Neubauer\inst{1} \and Johannes Scholz\inst{3} \and Radu Grosu\inst{2} \and Sophie Neubauer\inst{1,2}}
\authorrunning{J. Lemmel et al.}
%
\institute{Datenvorsprung GmbH \and
CPS Group, Technische Universität Wien\and
Technische Universität Graz\\$*$Equal Contribution}

\maketitle              
\begin{abstract}
Modern tourism in the 21st century is facing numerous challenges. Among these the rapidly growing number of tourists visiting space-limited regions like historical cities, museums and bottlenecks such as bridges is one of the biggest.
In this context, a proper and accurate prediction of tourism volume and tourism flow within a certain area is important and critical for visitor management tasks such as sustainable treatment of the environment and prevention of overcrowding.
Static flow control methods like conventional low-level controllers or limiting access to overcrowded venues could not solve the problem yet.
In this paper, we empirically evaluate the performance of state-of-the-art deep-learning methods such as RNNs, GNNs, and Transformers as well as the classic statistical ARIMA method. Granular limited data supplied by a tourism region is extended by exogenous data such as geolocation trajectories of individual tourists, weather and holidays.
In the field of visitor flow prediction with sparse data, we are thereby capable of increasing the accuracy of our predictions, incorporating modern input feature handling as well as mapping geolocation data on top of discrete POI data.

\end{abstract}

\section{Introduction}\label{introduction}

With increasing population and travel capacities (e.g. easy access to international flights) cultural tourism destinations have seen a rise in visitors. In addition, recent needs for social distancing and attendance limitations due to the global COVID-19 pandemic have confronted tourism destinations with significant challenges in e.g. creating and establishing sustainable treatment of the both urbanised and natural environment or e.g. preventing overcrowded waiting-lines. 
The perception of tourists regarding health hazards, safety and unpleasant tourism experiences may be influenced by social distance and better physical separation~\cite{SIGALA2020312}.

Based on The United Nation's 2030 Agenda for Sustainable Development~\cite{unwto}, tourism is obligated to contribute to several Sustainable Development Goals, including sustainable cities, responsible consumption, and economic growth. Sustainable tourism can achieve this by understanding and controlling visitor flows, preserving natural landmarks, reducing emissions and waste, establishing sustainable energy consumption, creating harmony between residents and tourists, and maximizing tourist satisfaction for economic prosperity.

Insufficient data availability in real-world problems is caused by factors such as compliance issues, lack of data collection, and transfer. Nonpersonal data from POIs, tourist facilities, and anonymized digital device data are used in research, but location data collected by mobile apps is controversial due to profit-oriented collection practices. It's important to consider whether people are aware of what they're sharing when using these services, even if the datasets don't contain direct personal data.
 
The question of how to improve awareness of data shared by such apps or services is not answered in this research.
This scientific work is focusing on what is possible to achieve in the given environment considering the given data and data history in regards to tourist flow prediction since sparse data is a widespread generic problem.

The first step in order to control tourist flows is to predict authentic movement and behavior patterns. However, since the tourist visitor flow is affected by many factors such as the weather, cultural events, holidays, and regional traffic and hotspots throughout a specific day, it is a very challenging task to accurately predict the future flow~\cite{DBLP:journals/corr/abs-1809-00101}. Due to the availability of large datasets and computational resources, deep neural networks became the state-of-the-art methods in the task of forecasting time-series data~\cite{ijcai2021-397}, including tourism flow applications~\cite{9271732}.

In this work, we focus on tourist flow prediction based on a local dataset from the visitors of the tourist attractions of the city of Salzburg as well as third-party geolocation data of individual tourists. After data preprocessing and dataset preparation, we attempt to compare the performance of different deep-learning-based methods for time-series prediction with ARIMA, a traditional statistics-based method. According to Li and Cao~\cite{liPredictionTourismFlow2018}, ARIMA is the most popular classical time forecasting method based on exponential smoothing and it was made popular in the 1970s when it was proposed by Ahmed and Cook~\cite{ahmedAnalysisfreewaytraffic1979} to be used for short-term freeway traffic predictions. 



We summarize the specific contributions of our paper as follows:
\begin{itemize}
    \item  We perform a comprehensive comparison of DL and ARIMA, a traditional technique, on a real-world dataset to reveal the shortcomings and point out necessary future improvements. 
    \item Per point-of-interest (POI), we perform granular predictions on an hourly basis, which is critical for the task of tourism flow control.
    \item We further evaluate modern DL techniques such as Transformers and GNNs.
    \item To the best of our knowledge, we are the first to apply a wide range of DL models to tourist flow prediction.
\end{itemize}

\section{Related Work}
\label{related work}
Considering the importance of predicting tourist flows in a growing industry, visitor forecasting has gained attention in recent years. Recurrent Neural Networks are used to forecast tourist demand, such as LSTMs that can be used in conjunction with deep neural networks or hidden Markov Models \cite{7905712,liPredictionTourismFlow2018}. Only a limited set of models is used in most of these studies to make predictions.

Another important aspect of tourism data is its granularity. Several studies focus on long-term estimates of monthly, quarterly, and yearly, or in the best case daily numbers of tourists in large regions as a measure of city or country-level tourism demand~\cite{kursor}. For tourism flow control, it is vital to perform granular predictions on an hourly basis and per POI. 

\textbf{DL-based models.} Time-series data prediction is typically handled by recurrent neural networks (RNNs). With RNNs, neural networks gain memory, allowing them to forecast sequence-based data. Gated RNNs are able to produce a good performance as LSTM~\cite{lstm} and GRU~\cite{GRU}. 

The RNN has limitations when faced with irregularly sampled time series, such as that encountered in tourist flow forecasting. In order to overcome this limitation, phased-LSTM~\cite{phased_LSTM} adds a time gate to the LSTM cells. GRU-D~\cite{GRUD} incorporates time intervals via a trainable decaying mechanism to deal with missing data and long-term dependence in time series.

instead of discrete-time models, continuous-time models with latent state defined at all times can also be used, such as CT-RNN~\cite{ctrnn}, CT-LSTM~\cite{CTLSTM}, and CT-GRU~\cite{CTGRU}, as well as NeuralODEs~\cite{ANODE}, which define the hidden state of the network as a solution to an ordinary differential equation. Augmented-NeuralODEs~\cite{ANODE} can alleviate some limitations of NeuralODEs, such as non-intersecting trajectories, by using augmentation strategies. These continuous-time models have favorable properties, such as adaptive computation and training with constant memory cost. GoTube~\cite{GoTube2022} can be used to statistically verify them by constructing stochastic reach tubes of continuous-time systems.

On the other hand, transformer-based models~\cite{vaswani2017attention} have been successful in various applications due to their powerful capability for sequence learning and representation. They have also been explored in time-series forecasting tasks for datasets with long sequences and high historical information. The multi-head self-attention mechanism is the primary component of transformer models, which can extract correlations in long sequences. However, the permutation-invariant nature of self-attention requires positional encodings to prevent the loss of temporal dependencies.

Graph Neural Networks (GNNs) are an interesting new class of Deep Learning Algorithms that allow for the inputs to be structured as graphs. Most GNN models build on the notion of Graph Convolutions which can be seen as a generalization of Convolutional Neural Networks to graph structured data - as opposed to being arranged in a grid. An even more fascinating type of DL models are temporal GNNs that combine Graph Convolutions with RNNs. Such temporal GNN models are most prominent in traffic flow prediction applications. \cite{zhu2020}

\textbf{Traditional techniques.} For time-series forecasting with traditional techniques we use the Autoregressive Integrated Moving Average (ARIMA) model.
ARIMA has been used in recent studies as a baseline for the evaluation of novel deep-learning based models~\cite{yaoNeuralNetworkEnhanced2020} and is thus selected as a baseline model for this paper as well.

\section{Data}\label{sec:data}

Two different data sources were combined to enable the use of their different features in the training of the models and prediction of future visitor counts.  

The first dataset we used stems from the "Salzburg Card" which was kindly provided to us by TSG Tourismus Salzburg GmbH. Upon purchase of these cards, the owner has the ability to enter 32 different tourist attractions and museums included in the portfolio of the Salzburg Card. The dataset consists of the time-stamps of entries to each POI. Additionally, we used data about weather and holidays in Austria. 

We utilized mobile phone location data from a third-party service to improve tourist flow predictions in Salzburg. The dataset covers around 3\% of tourists and provides information on the number of tourists between points of interest. However, the data is sparse and lacks a distinct recording frequency. To further improve our predictions, we incorporated a street graph obtained from OpenStreetMap using the \texttt{osmnx} python package. The resulting graph contains 2064 nodes and 5359 edges, with edge values corresponding to the lengths of the street segments. We then mapped the location data to the graph by assigning each location to the nearest node and aggregating the total number per hour.


\paragraph{CoVID-19}
Tourism around the globe saw huge drops during the global CoVID-19 pandemic. Starting in march of 2020, Austria started to take preemptive measures to prevent the spread of the virus. These travel restrictions and closings of public spaces, hotels and restaurants severely reduced the number of tourists in and around the city of Salzburg. As a consequence, prediction accuracy could be diminished when using models that have been trained on pre-CoVID data.

\section{Methods}
For this work, we built our own dataset on hourly data collected from tourist attractions and then expanded this by including geolocation data. Including many different datasources is a key challenge for this real-world prediction task. Sparse geolocation data is therefore fed into our GNN model as features. With this approach we aim to create models that are capable of easily integrating new datasources that might be available in the future. 
We then perform predictions with a rich set of models and do a comprehensive comparison of the results. 
In this section we first introduce the dataset we used for the experiments. Then we go over the methods we chose to evaluate and compare their performances. 

\subsection{Deep-Learning models}
We use a large set of RNN variations on the tourist-flow dataset to perform a comprehensive comparison of the state-of-the art models and provide insight on their performance. The set comprises vanilla-RNN, LSTM, phased-LSTM, GRU-D, CT-RNN, CT-LSTM and Neural-ODE networks. Moreover, we used a Transformer model, using only the encoder part with 8 heads, 64 hidden units, and 3 layers, to forecast the tourist flow. Finally, we applied a naive continuous-time temporal GNN approach based on CT-RNNs to our prediction problem in order to utilize geolocation data of individual tourists. All of the Neural Networks were trained with Backpropagation-Through-Time and the Adam optimizer \cite{kingma2014method} using the parameters given in the Appendix \ref{table:Hyperparameters}. 

In order to incorporate inductive bias stemming from the street layout from Salzburg, we used a simplified CT-RNN based GNN model that we will call Continuous-Time Recurrent Graph Network (\textbf{CT-GRN}) in the following. It consists of one neuron per node in the street graph and exhibits the same connectivity. This is done by point-wise multiplying the recurrent kernel with the graph's normalized adjacency matrix whose entries are the inverse of the corresponding street segment lengths.
$$y_{t+1} = y_t -\tau y_t + a \odot \tanh((W_{rec} \odot \hat A) y_t + W_{in} x_t + b)$$

Where $y_t$ is the network's state at time $t$, $x_t$ is the exogenous input, $\tau$, $a$, $W_{rec}$, $W_{in}$ and $b$ are trainable parameters, and $\hat A = D^{-1}A$ is the normalized Adjacency matrix. The resulting model inherits all the favourable ODE properties of CT-RNNs such as the ability to evaluate at arbitrary points in (continuous) time and differentiable dynamics used in verification. Finally, we used a variation of the \textit{Teacher Forcing} \cite{williams1989} technique which basically translates to resetting the nodes of the network to the target value after each step. Our \textit{\textbf{Mixed} Teacher Forcing} version forces the hidden state of the POI nodes to the true value and adds up the predicted and true values for the other nodes.

\subsection{Traditional methods}

In this study, we used a non-seasonal ARIMA model (\emph{ARIMA ($p$,$d$,$q$)}) that ignores seasonal patterns in a time-series, where $p$ is the number of autoregressive terms, $d$ is the number of non-seasonal differences, and $q$ is the number of lagged forecast errors in the prediction equation~\cite{burgerPractitionersGuideTimeseries2001}. We utilized the \emph{auto.arima} function from the \emph{R forecast} library to automatically determine the best values for $p$, $d$, and $q$ for each of the 32 POIs. The ARIMA model was then individually fitted to each POI's training dataset using the \emph{pmdarima} library in Python. Each time the number of visitors is predicted for the next hour in the test data, the true value (i.e., number of visitors) for that hour is added to update the existing ARIMA model and make it aware of all previous values before making the next prediction. This approach prioritized prediction accuracy over time complexity.

\subsection{Preprocessing}
We used the Salzburg card data from years 2017, 2018, and 2019 for our first set of experiments. In order to create the time-series data, we accumulated the hourly entries to each location. The data then consists of the hour of the day, and the number of entries at that hour to each of the 32 POIs. 

For the DL models, we added additional features to the dataset: Year, Month, Day of month, Day of week, Holidays and Weather data. For the Holiday data we used the national holidays and school holidays and count the days to the next school day. For the Weather data, we used the hourly weather data with these features: Temperature, Feels Like, Wind speed, Precipitation, and Clouds as well as a One-Hot-Encoded single word description of the weather (e.g. "Snow").

We performed further pre-processing by normalizing all features to values between $0$ and $1$. To account for seasons, we performed sine-cosine transformation for the month. Intuitively, since it is a circular feature we do not want to have the values for December and January to be far apart. 

Finally, we split the data into sequences of length 30, and used the data from years 2017 and 2018 as the training set, and 2019 as the test set.

\begin{figure*}
    \centering
    \includegraphics[width=0.8\textwidth]{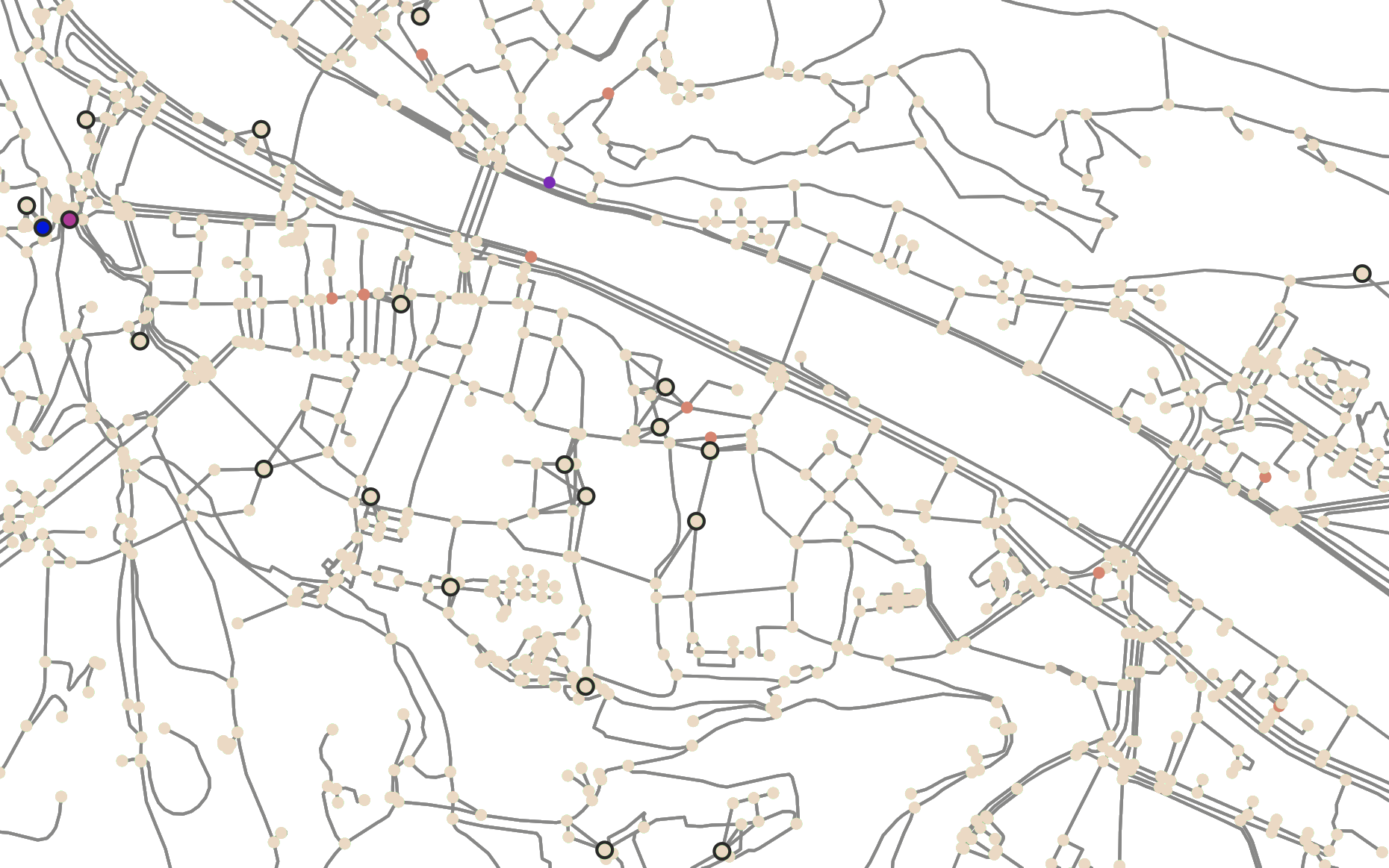}
    \caption{One sample of the series of OSM graphs of the Salzburg city center obtained from preprocessing. Encircled nodes are the special POI nodes. Color coded are the normalized aggregated entry and tracking data, where most of the nodes indicate zero (pale)}
    \label{fig:graph}
\end{figure*}

\textbf{Graph Neural Networks} For the GNN we used the OSM graphs as illustrated in Section \ref{sec:data}. Our dataset of tourist locations was very sparse which subsequently resulted in very sparse inputs for each node. Since we are trying to predict numbers of entries at the POIs, we added them as additional nodes to the graph connecting them to up to 5 of the nearest nodes present in the graph with a max distance of 80 m. Finally, the global features such as weather and holidays are added to the graph by a linear mapping from features to nodes. This way we obtained a series of graphs where each sample constitutes the OSM graph with the edge values corresponding to the distance and the node values corresponding to the aggregated number of people near this location / POI entries. One sample is visualized in Figure \ref{fig:graph}. For inference we predicted the whole graph and discarded the nodes that do not represent POIs.

\begin{table*}[ht]
  \centering
  \caption{Averaged prediction errors}
  \begin{tabular}{|l|c|cc|cc|cc|}
    \hline
        & \# Cells / & \multicolumn{2}{c}{Time} & \multicolumn{2}{|c|}{only visitors}&\multicolumn{2}{c}{external features}\\
    \textbf{Model}     &  \# Parameters & Train (min) & Pred (ms) & MAE & RMSE & MAE & RMSE \\

    \hline
    ARIMA &  \textbf{224} & - & 69k & 5.217 & 7.833 & - & -\\
    \hline
    ANODE           &    64 /       21.3k     &       145.6   & 3.01 & 4.599 & 6.965 &         4.410   &       6.663   \\
    Vanilla RNN     &   128 /       43.7k     &       5.9     & \textbf{0.18} & 3.958 & 6.321 &         3.802   &       6.160   \\
    LSTM            &    32 /       11.9k     & \textbf{1.5}  & 0.24 & 3.713 & 6.209 &         3.630   &       6.113   \\
    Phased LSTM     &    32 / \textbf{11.8k}  &       27.0    & 0.46 & 3.825 & 6.359 &         3.651   &       6.120   \\
    CT-LSTM         &    32 /       19.9k     &       18.1    & 0.31 & 3.734 & 6.239 &         3.700   &       6.185   \\
    CT-RNN          &   128 /       27.4k     &       57.1    & 0.60 & 3.694 & 6.131 &         3.629   & \textbf{5.983} \\
    GRU-D           &    64 /       27.7k     &       16.6    & 0.33 & \textbf{3.638} & \textbf{6.121} &  \textbf{3.621} &       6.073   \\
    \hline
    Naive & - & - & - &  6.466 & 9.483 & - & - \\

    \hline
  \end{tabular}
  \label{table:results}
\end{table*}



\section{Main Results}

\subsection{Forecasting visitor numbers}

We performed a diverse set of experiments with ARIMA and DL models to evaluate their forecasting accuracy, execution time and prediction time and compare the models. Table~\ref{table:results} shows the Mean-Absolute-Error (MAE) and Root-Mean-Squared-Error (RMSE) achieved for each method applied to the timeframe from 2017-2019 - before COVID. In order to find optimal model size, loss function, and whether to use normalized visitor counts, we did a grid search conducting three runs per configuration and keeping the one which achieved the lowest average RMSE. As a baseline we include the naive approach of using the last true value as prediction at each step, i.e. $\hat y_t = y_{t-1}$.

The table includes the model size, number of parameters, and training and prediction times for the best run of each deep-learning model. We excluded non-normalized models since normalized visitor counts consistently led to better results. MAE was the best loss function for all models except ANODE, which performed better with Huber loss. The phased LSTM achieved comparable results with the fewest parameters.

Our DL models outperformed ARIMA in both metrics, with and without additional features. Adding more features did not significantly improve performance, suggesting that it may lead to over-fitting. We report results with and without additional features for DL models to ensure fairness in comparison with ARIMA, which cannot use external features. Additionally, ARIMA struggles with short sequences, while DL models can handle them when trained on the full dataset.


\begin{figure*}[ht]
    \centering
    \includegraphics[width=0.75\textwidth]{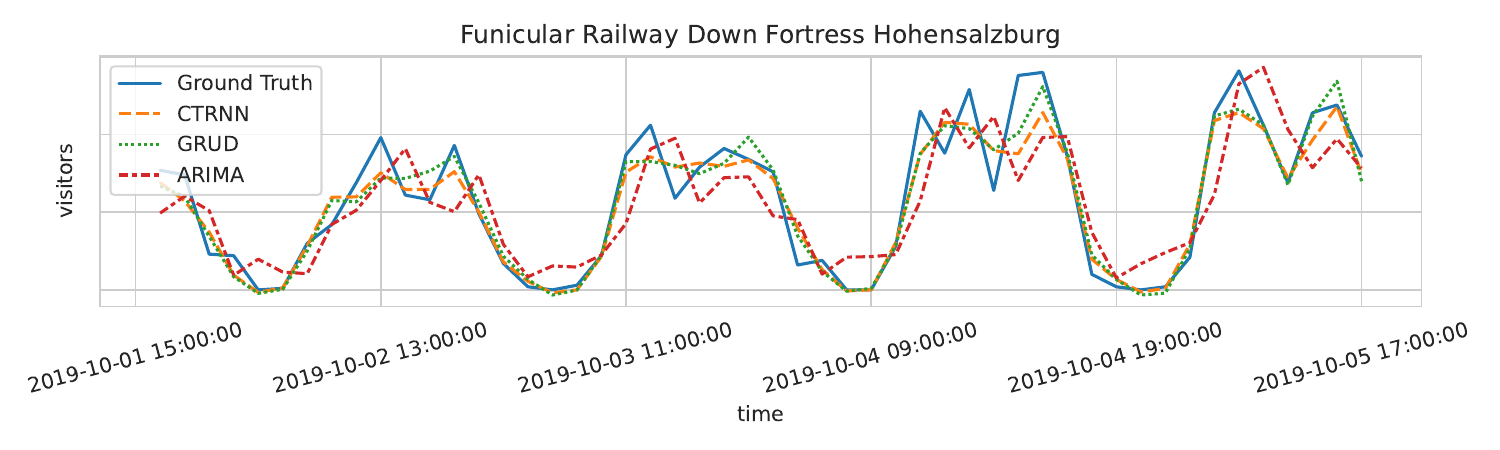}
    \includegraphics[width=0.75\textwidth]{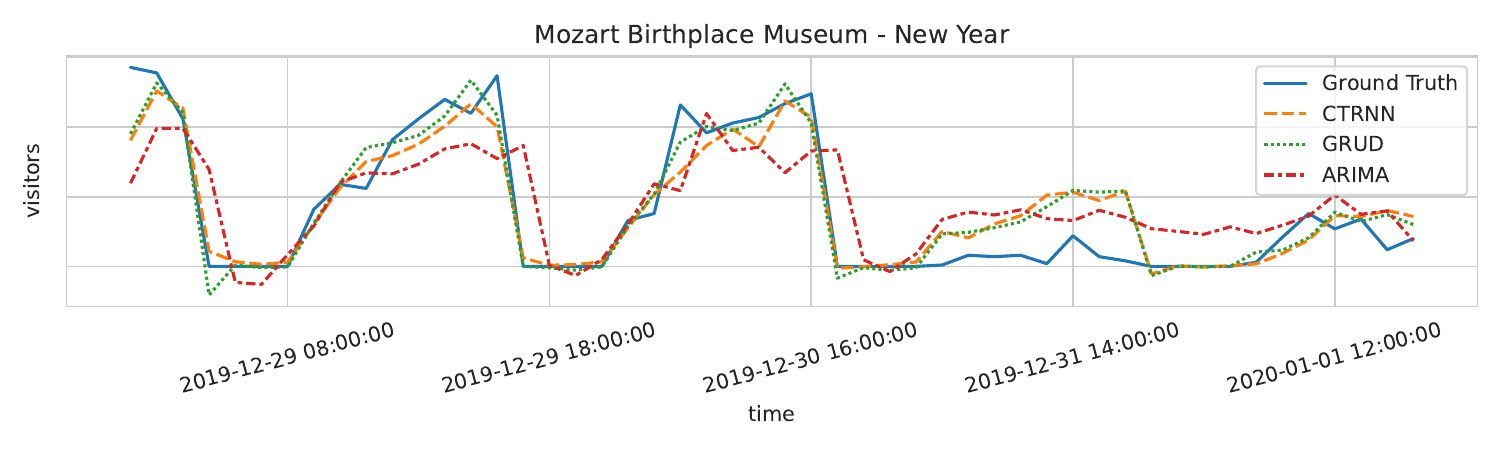}
    \includegraphics[width=0.75\textwidth]{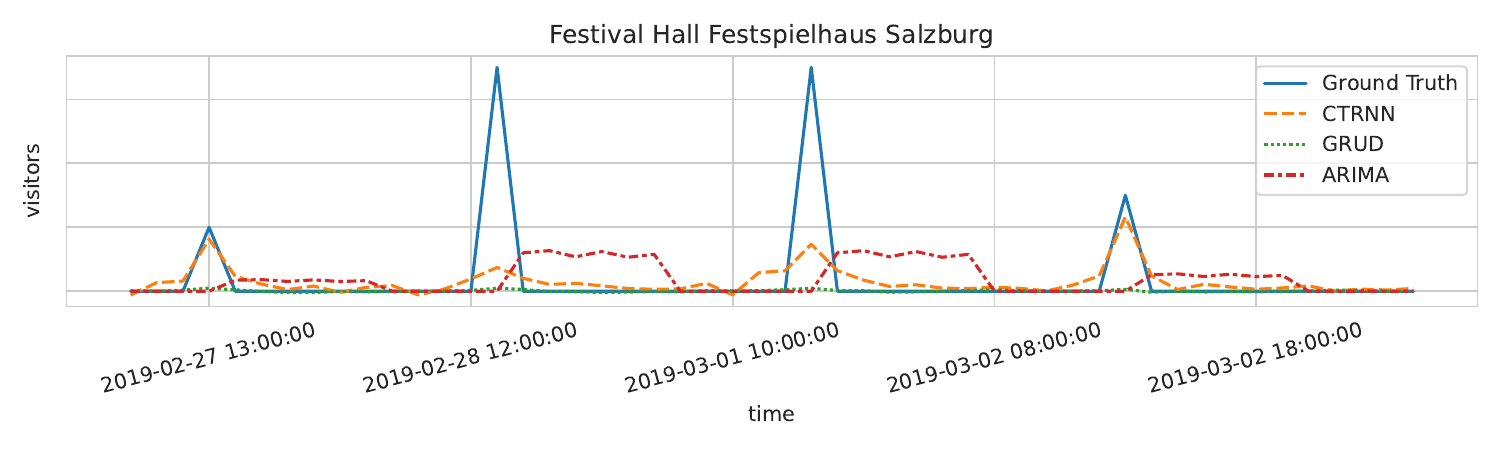} 
    \caption{Predicted and True visitor counts for the Funicular Railway (top) and Mozart's Birthplace Museum (mid) and the Festival Hall (bottom). Predictions are computed using CT-RNN (orange), GRU-D (green) and ARIMA (red).}
    \label{fig:predictions}
\end{figure*}

In Table~\ref{table:results}, we compare the training and prediction times of ARIMA and DL models. ARIMA took 69s to perform a single prediction for all POIs, while the DL models took fractions of milliseconds, with the trade-off of having longer training times. ARIMA does not have a dedicated training step, and its calculations are time-consuming since it makes predictions for each POI separately. In contrast, DL models are trained with the visitors to all POIs in a single vector and make predictions for all at the same time. This allows DL models to leverage implicit data about the total number of visitors in the city, which ARIMA loses.

    
In order to visually explore the predictions made by the models, we plotted the predictions and the ground truth for a few selected time-windows (see Figure \ref{fig:predictions}). We plot the predictions made by the DL models (including the external features) with the best MAE and RMSE, which were the GRU-D and CT-RNN respectively. The prediction made by the DL models with the visitors only data was only slightly worse than the others, which is why we omit these evaluations in the plots. Our plots show that although ARIMA is out-performed by the DL methods in the average error of all predictions, there are cases where it actually performs better than the other models. 
The plot on the Top shows the forecast and real values for the tourists entered the Funicular Railway descend which is the cable car ride leading up to Salzburg Castle. As visible in the plot, the DL models show a better performance, especially in the valleys where the ARIMA fails to predict the downfalls accurately. 
Mid shows visitor predictions for Mozart's Birthplace Museum around the time of New Year's Eve. The reduced numbers of visitors on the 1st and 2nd of January is overestimated by all our models.
Finally, on the bottom the predictions for the Festival Hall Festspielhaus guided tour are shown which is sparse since it takes place once a day at 2 pm. All models fail in prediction for the second and third peak at this location. However, CT-RNN shows a very good performance in predicting the first and last peak and at least shows an upward trend for the second and third peak. ARIMA can not handle this type of sparse data at all.

\begin{table}[ht]
  \centering
  \caption{Average prediction results for 2021 after training on data from 2019 \& 2020}
  \begin{tabular}{|l|c|c|c|}
    \hline
          &     \multicolumn{3}{c}{MAE}   \\
          
    \textbf{Model}  & only visitors& + features & + geolocation\\

    \hline
    Vanilla RNN    &  2.48 &  2.42  & 3.37 \\
    LSTM           &  2.66 &  2.58  & 3.54 \\
    Phased LSTM    &  2.44 &  2.44  & 3.05 \\
    CT-LSTM        &  2.61 &  2.57  & 3.32 \\
    CT-RNN         &  2.50 &  2.45  & 3.16 \\
    ANODE          &  2.63 &  2.57  & 3.64 \\
    GRU-D          &  2.99 &  2.87  & 3.58 \\
    Transformer    &  2.19 &  \textbf{2.04} & 2.65 \\
    \hline
    Naive & 2.187 &  - & - \\
    \hline
    CT-GRN         &  - & - & \textbf{2.63}\\
    \hline

  \end{tabular}
  
  \label{table:results2}
\end{table}

\subsection{Including geolocation data}
We conducted a second set of experiments on the timeframe from 2019 to 2021 that includes geolocation data of individual (anonymized) tourists. Results are presented in Table \ref{table:results2} which shows for each model the number of Parameters and MAE with and without additional features and also when using features and the geolocation data. This time we included the Transformer and GNN models, but excluded ARIMA due to computation time reasons. Since the \textit{Salzburg Card dataset} for this particular timeframe contains a significantly lower number of datapoints due to lockdowns enforced by the government, the numbers must not be compared directly to the results discussed in the last section. 

This time the naive approach outlined above led to surprisingly good results and only the Transformers with exogenous features were able to surpass it. Transformers can handle multi-variate data well due to the multi-head self-attention mechanism which enables them to extract hidden correlations in input, and hence get better loss after using additional features. However, they require considerably more parameters in comparison to the RNN models. 

For the GNN we only conducted experiments with additional geolocation data since input graph attributes would be even sparser without, defeating the point of using a graph based approach. The CT-GRN algorithm scored a slightly worse prediction error in comparison to the other models when not using visitors and features as input. However, all other methods scored worse when trained on the sparse geolocation data which shows the usefulness of the GNN approach.

Our GNN approach enables us to incorporate the sparse geolocation data into our model. Since there is more sparse geolocation data expected to be processed within real-life-scenarios, this is the only approach to fit these needs. 

\section{Conclusions and future work}

Our study demonstrated the effectiveness of DL models for tourist flow time-series forecasting, particularly when external features are included. DL models outperformed the traditional ARIMA method and were faster in terms of prediction time. We also showed that GNNs are more suitable for incorporating spatial structure using sparse geolocation data.

Moving forward, there are several directions for future research. One possibility is to investigate methods to further improve the performance of DL models, such as implementing regularization or learning rate scheduling. Another option is to explore the use of Vector Auto-Regression (VAR) to address the limitations of ARIMA for univariate data. Finally, we plan to develop specialized models that can outperform existing state-of-the-art models in short-term prediction, with the ultimate goal of helping tourism stakeholders make informed decisions and promote sustainable tourism practices.

\section*{Acknowledgements}
This work is supported by the Austrian Research Promotion Agency (FFG) Project grant No. FO999887513. SG is partially funded by the Austrian Science Fund (FWF) project number W1255-N23. Map data copyrighted OpenStreetMap contributors and available from \url{https://www.openstreetmap.org}.

\nocite{*}
\bibliographystyle{splncs04}
\bibliography{conference_idsc}

\appendix
\section{Appendix}


\subsection{Hyperparameters}
\begin{table}[ht]
    \centering
    \caption{Hyperparameters used in RNN training.}
    \begin{tabular}{ |l c| }
    \hline
     \textbf{Hyperparameter} & \textbf{Value} \\
     \hline
     sequence length & 30\\
     batch size & 16\\
     epochs & 300\\
     \hline
     optimizer & adam \\
     Learning-rate & $1e^{-3}$ \\  
     $\beta_{1,2}$ & $(0.9, 0.999)$\\    
     $\epsilon$ & $1e^{-8}$ \\
     \hline
     loss function & mse, mae, huber\\
     model size & 32, 64, 128\\
     normalized visitors & True, False\\
     \hline
    \end{tabular}
    \label{table:Hyperparameters}
\end{table}

%
%
%
%
\end{document}